\documentstyle[epsf]{article}
\begin{document}
\noindent\begin{picture}(0,0)
\put(0,0){\parbox{\textwidth}{%
Appeared in {\em International Journal of Approximate\\ 
Reasoning}, 19(1--2):161--191, July/August 1998.}}
\end{picture}
\thispagestyle{empty}
\mbox{}\\[1cm]
\begin{center}
\fbox{
\begin{minipage}{11cm}
\begin{center}\mbox{}\\*[4mm]
{\Large {\bf A reusable iterative optimization software library to solve
combinatorial problems with approximate reasoning}}\\*[7mm]
{\bf Andreas Raggl \quad Wolfgang Slany}\\*[4mm]
Institut f\"{u}r Informationssysteme E184-2\\*
Technische Universit\"{a}t Wien, A-1040 Wien, Austria\\*
mailto:wsi@dbai.tuwien.ac.at\\*
http://www.dbai.tuwien.ac.at/staff/slany/\\*[1mm]\mbox{}
\end{center}
\end{minipage}}\\\vfill\vfill
{\Large {\bf DBAI-TR-98-23}}\\\vfill\vfill
\mbox{}\hfill\epsfysize=2cm\epsfbox{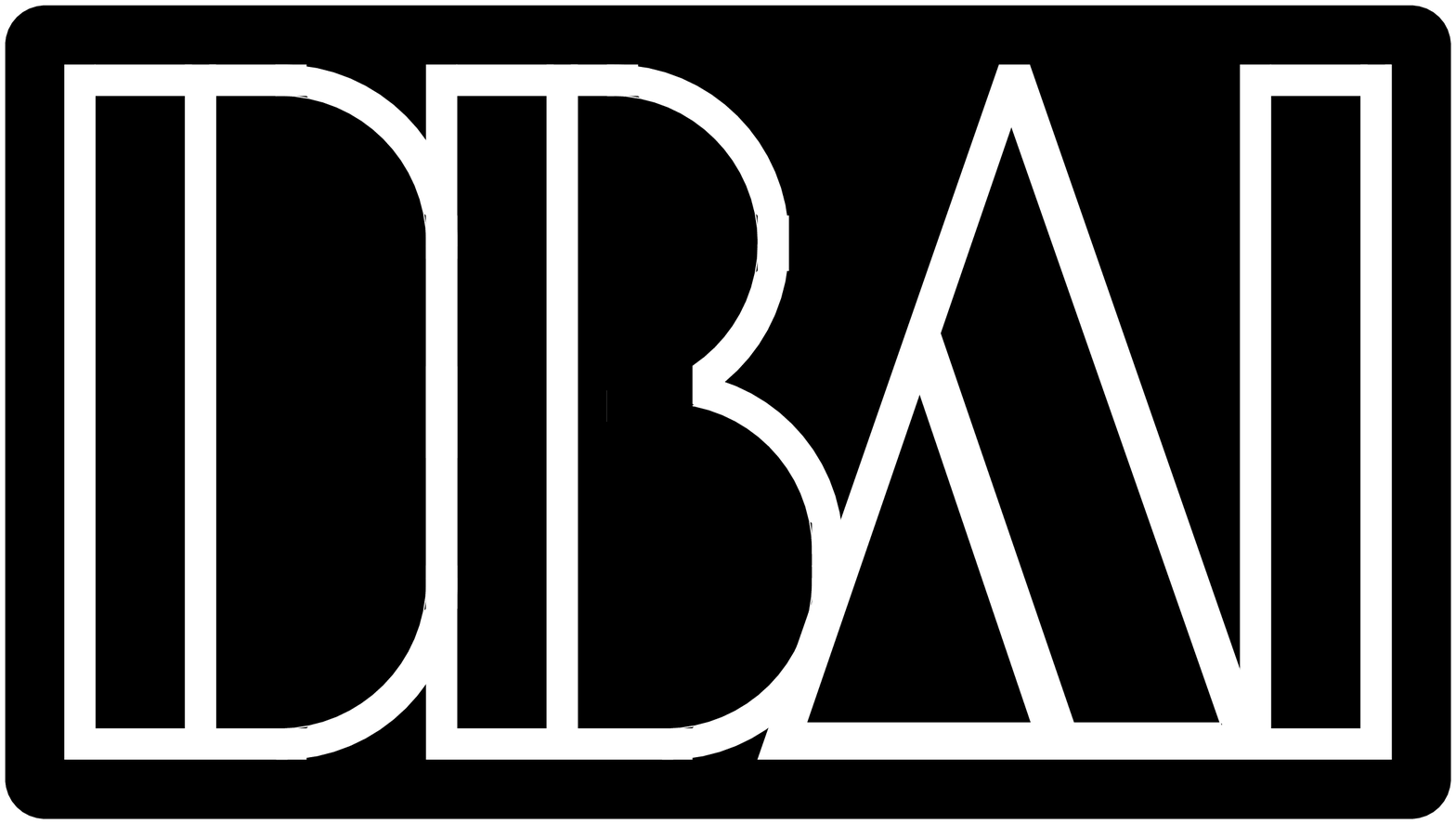}\hfill\mbox{}\\\vfill
    {\bf Abteilung f{\"u}r Datenbanken und Expertensysteme\\*
      Institut f{\"u}r Informationssysteme\\*
      Technische Universit{\"a}t Wien}\\\vfill
\mbox{}\hfill\epsfbox{tu-logo.eps}\hfill\mbox{}
\end{center}
\newpage\pagenumbering{arabic}

\def\epsfsize#1#2{\hsize}
\title{A reusable iterative optimization software library to solve
combinatorial problems with approximate reasoning}
\author{Andreas Raggl\\[1ex]
       Securities \& Treasury\\
       Erste Bank, A-1010 Wien, Austria\\
       Email: raggl@erste.at\\[3ex]
       Wolfgang Slany\\[1ex]
       Institut f\"{u}r Informationssysteme E184-2\\
       Technische Universit\"{a}t Wien, A-1040 Wien, Austria\\
       Email: wsi@dbai.tuwien.ac.at}
\date{}
\thispagestyle{empty}
\maketitle

{\abstract Real world combinatorial optimization problems such as
scheduling are typically too complex to solve with exact
methods. Additionally, the problems often have to observe vaguely
specified constraints of different importance, the available data may
be uncertain, and compromises between antagonistic criteria may be
necessary. We present a combination of approximate reasoning based
constraints and iterative optimization based heuristics that help to
model and solve such problems in a framework of {\em C++\/} software
libraries called StarFLIP++. While initially developed to schedule
continuous caster units in steel plants, we present in this paper
results from reusing the library components in a shift scheduling
system for the workforce of an industrial production plant.}

\mbox{}

\newpage
\tableofcontents

\section{Introduction}

Government as well as industry require practical approaches to a
diverse set of complex combinatorial optimization problems.  In
industry, the distinction between commercial viability and failure
often lies in the ability to control the production process through
efficient optimization.  Scheduling is one example of such
combinatorial optimization problems. Like most combinatorial
optimization problems of practical relevance, it is usually very hard
to solve, both in practice as well as for theoretical reasons. Results
from complexity theory \cite{Lawler93} indicate that in the worst
case, the fastest algorithm that is able to find the optimal solution
of a typical problem can only be as fast as an algorithm that compares
all possible schedules. Since the search space is by far too big,
systematic search must be ruled out, and it therefore seems clear that
some random sampling technique has no worse chance to hit relatively
`good' solutions than any other algorithm.  While scheduling has been
studied in isolation for many years, recent advances in artificial
intelligence and operations research indicate a renewed interest in
the area \cite{Zweben94}. In addition, the scheduling problem is being
defined more generally, and work is beginning to consider the closed
loop use of scheduling systems in operational contexts. However, a
primary source of difficulty in constructing good schedules stems from
the conflicting nature of the objectives.

As with many real life decision making situations, it is usually not
possible to fulfill perfectly all objectives when building new
schedules.  This applies to classroom schedules, staff rosters, as
well as production schedules in manufacturing. Existing approaches to
scheduling have tended to reduce the complexity of the problem by
considering only a {\em small subset\/} of objectives. In real world
situations, it would often be more realistic to find {\em viable
compromises\/} between the objectives. For many problems, it makes
sense to partially satisfy objectives. The satisfaction degree can
then be used to evaluate the achieved compromise. In addition, real
objectives are often prioritized, therefore it is necessary to weight
their satisfaction with importance factors. One especially
straightforward way to achieve these two aspects of scheduling
problems --- to satisfy constraints to a certain degree, and to take
into account relative importances --- is the modeling of these
constraints through fuzzy constraints. Fuzzy constraints are
particularly well suited for modeling, since constraints can be
written in a format easily understood by human experts, and because
they feature a robust behavior which needs almost no tuning to yield
reasonable control. In addition, the evaluation of their gradual
satisfaction can be very efficiently used to guide repair based
heuristic search methods as described for instance by Slany in
\cite{Slany96a}, in order to find approximate `good' solutions while
at the same time greatly reducing the time needed to find them.

Repair based heuristic search methods are local methods that collect
information on the problem by more or less random sampling it at
various points, and mainly differ in the way the next random sample is
chosen. A step from one sample to the next is defined by a
neighborhood concept. Functionally, this neighborhood concept is
implicitly defined through so-called repair operators that represent a
transition from one variable instantiation to another one, both
corresponding to more or less possible schedules. Repairing a random
initial and typically bad schedule therefore corresponds to applying a
series of repair operators until one reaches a neighborhood in which
the included schedules violate few constraints, and thus get better
evaluation scores than the random initial one.

These repair based heuristic search methods stand in contrast to the
more systematic, traditional constructive methods. There, a feasible
schedule is built from scratch, i.e., the variables initially are all
uninitialized and step by step are assigned values by the
algorithm. If a deadlock is reached, some variables that had already a
value assigned must be reinitialized and a new search path has to be
chosen. In practice, there is a plethora of different methods that
basically follow this line of thought: Common to them all is that
they in principle do not work on complete instances that still violate
some constraints, but instead build-up the schedule constructively.

Additionally, since fuzzy constraints allow a wide range of values for
variables, the constructive approaches are faced with an even huger
search space compared to the usual constraint problems. By intuition,
this huge search space lends itself in a much more natural way to
random sampling techniques such as repair based methods. On the other
hand, mathematical analysis is made much more difficult in the random
sampling case combined with multi-criteria non-linear fuzzy
constraints.  However, empirical benchmark results indicate that the
aforementioned intuition is right, in that the performance of repair
based heuristic search methods on real world problems is much better
than the performance of constructive methods, see \cite{Slany96a,
Dorn96}.

Real world descriptions naturally contain vaguely formulated
relations, because further details are simply not known or would
anyway not lead to better results as they would be canceled out
through noise in the data.  The down-to-earth reason behind our choice
of fuzzy logic as a basis for knowledge representation is that it
allows straightforward modeling of typical combinatorial optimization
problems and is perfectly combinable with heuristics that find `good'
solutions in acceptable time.

Repair based heuristics have a much better efficiency to solve typical
large optimization problems compared to constructive or enumerative
algorithms. In particular, they need no explicit constraint relaxation
to still be able to implicitly assess trade offs between conflicting
constraints when the latter are modeled using the mentioned fuzzy
constraints. Further, these repair based heuristics do not need to
prune search space to still yield very good results for well-known
benchmark problems. Indeed, almost all other fuzzy constraint
satisfaction algorithms found in the literature (see \cite{Slany96a}
for a survey) rely on search space pruning to achieve better
performance, but often explicitly do not look at possibly better
compromise solutions (in particular methods that prune all paths where
$\alpha$-cuts fall below a certain level), implying that a solution
featuring a relatively unimportant sub-constraint with very low
satisfaction but constituting nevertheless the real optimum because of
the other, more important constraints being satisfied to a higher
degree than in all other instantiations, could be neglected forever.
In this sense, the method proposed in the StarFLIP++ project could be
seen as an --- albeit not 100\% perfect --- solution to the question
whether fuzzy set theory can solve large and complex problems
computationally efficiently.

Additionally, in industrial applications, reacting to a changing
situation, i.e., rescheduling has to be done quite frequently when
some production parameters change due to machine breakdowns. Usually,
most human errors are made in these rescheduling situations since time
to think is scarce and the situation often worsens rapidly (e.g.,
forgetting for some time a waiting machine, resulting in longer
waiting times or worse qualities for certain jobs) if no action is
taken. Iterative optimization based methods are inherently well suited
to deal with such situations.

The paper is organized as follows. The next section introduces the
shift scheduling application. Section~\ref{section:starflip} then goes
on to present the major components of the StarFLIP++
project. Section~\ref{section:shiftcons} presents the constraints and
repair steps of the shift scheduling application that we chose to
model with StarFLIP++. Section~\ref{section:stats} presents specific
details of the challenges encountered in integrating the concepts to
the shift scheduling problem in the StarFLIP++ framework and presents
benchmark results indicating the effectiveness of StarFLIP++ for this
kind of combinatorial optimization problem. Finally, we conclude and
take a look at possible future steps that will make StarFLIP++ even
more useful in a distributed context on the Internet.


\section {The Shift Scheduling Application}\label{section:shiftpb}

Right from its beginning the StarFLIP++ project has always been
strongly coupled with problems encountered in the process of steel
production. This was partly due to the fact that the entire project
has been initiated by a research cooperation with the Austrian steel
production industry. A wide range of publications have been published
on the steel production domain over the last couple of years out of
this fruitful research cooperation. Slany gives in~\cite{Slany96a} a
more in-depth discussion of this domain in connection with fuzzy
scheduling. Dorn and Shams~\cite{Dorn91h} discuss an expert system
approach designed initially for a similar domain.

With versatility and reuse being key objectives of the StarFLIP++
project, we chose to move on from the original steel production
domain. We expected to gain further experience about the process of
knowledge acquisition and transformation into a StarFLIP++ compatible
format, which led us eventually to a system that is more or less a
generic tool as far as the representation of domain knowledge is
concerned. Secondly, another problem domain also gave us numerous
hints on weaknesses of the system. These weaknesses were located in
the optimization methods, in the performance of the system, and in the
representational power provided by the fuzzy tools of the FLIP++
library.

The shift scheduling domain is a promising field of application for
several reasons. To begin with, scheduling research in this area is
almost nonexistent despite the fact that it is an important but
difficult application area. One major conclusion drawn out of the
existing research efforts is the fact that one soon runs into major
difficulties in this area when conventional optimization methods are
applied, e.g. with simplex, enumeration or backtracking methods.
According to G{\"a}rtner and Wahl~\cite{Gaertner95a} a high degree of
fuzziness can be attributed to many requirements encountered with
shift scheduling problems. They also argue that due to the complexity
of the problem and the lack of powerful optimization methods it is
more important to move the focus from automation of design towards
aiding design. We also believe that any system used to solve such
shift scheduling problems must be a cooperative tool that allows to
find an optimal schedule via the interaction with the knowledge
engineer. Nevertheless, StarFLIP++ contains elements that allow to use
it eventually as a closed loop system. Moreover, the flexibility
offered by StarFLIP++ when it comes to the definition of fuzzy
variables, fuzzy constraints and aggregation operators should make it
superior compared to classical optimization methods which often show a
lack of representational power. Moreover, the repair based
optimization process that tries to tackle constraint violations with
specifically defined repair steps shows very good results as discussed
in Section~\ref{section:stats}.

Furthermore, the problem of developing `good' shift schedules is a
highly practical application. It has many consequences on people that
are working in shifts. The industrial optimization potential and
social implications of shift schedules (e.g. consequences on family
life) are considerable.

In the present paper, we describe a subset of the actual constraints
in order to focus on the major aspects of StarFLIP++, the shift
scheduling application serving only as an illustration to the program
description. In particular, the number of represented constraints was
reduced by focusing on a problem instance with simple shift types. For
example, the concept of night shifts has been completely left out.
Consequently, all constraints referring to night shifts could be left
out. However, the example was chosen with sufficient complexity to
illustrate the main points of StarFLIP++. Once a proper representation
of a problem is found, enhancing the constraints of the problem does
not cause much difficulty.

The objective of our problem is to find a shift schedule for twelve
employees.  These employees are aggregated in three groups with an
even distribution, i.e., each group consists of four employees. The
groups are further divided into subgroups of two employees each. Each
of the subgroups is fully covering the requirements of operation,
hence no interdependencies between the various subgroups have to be
taken into consideration. Ruling out interdependencies is a further
simplification that is rare in real world problems but makes it
easier to follow the problem description. Again, as mentioned above,
such a simplification does not impede the judgment of the basic
functionality of StarFLIP++ in connection with a shift scheduling
problem, as there will be enough constraints to allow a rich and highly
nonlinear interaction. The groups are named {\em A}, {\em B}, and
{\em C\/} with subgroups {\em A1}, {\em A2}, {\em B1}, {\em B2}, {\em
C1}, and {\em C2}, respectively. The working hours are 38.5 hours per
week.  Weekly working hours can vary over the length of the shift
schedule, but the average per week must be 38.5 hours. There exist
several shift types that can be allocated only at specific times (see
Figure~\ref{figure:shiftdefinitions}).

\begin{figure}[th]
\begin{center}
\begin{tabular}{|l|r|l|}\hline
name & length (in hours) & shortcut\\ \hline
day shift: 			& 8--9 	& (TD)\\ \hline
day shift at weekends: 		& 4 	& (TDWE)\\ \hline
shift substitution at weekends: & 12 	& (SWWE)\\ \hline
\end{tabular}
\end{center}
\caption{Shift type definitions.}
\label{figure:shiftdefinitions}
\end{figure}

\begin{figure}[th]
\epsfbox{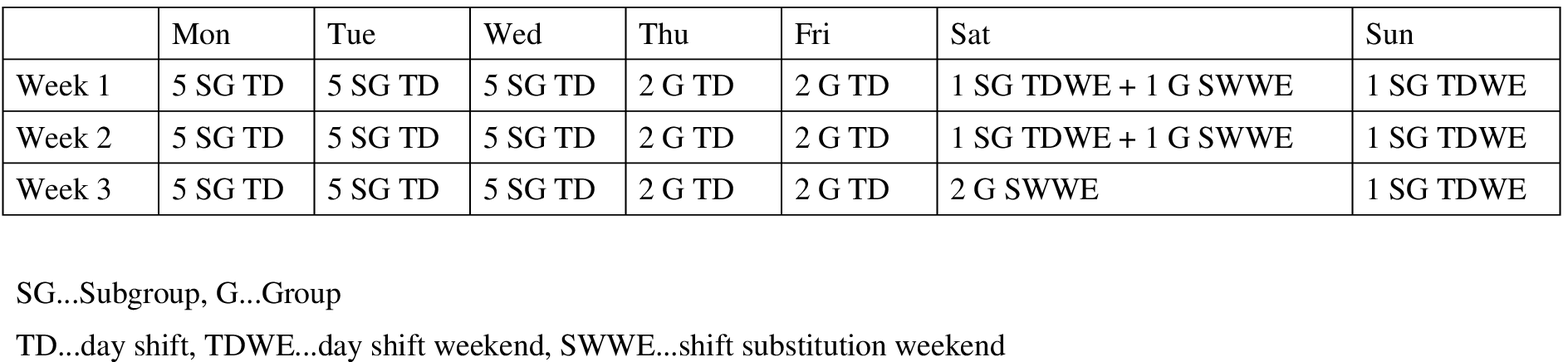}
\caption{Operation plan.}
\label{figure:operationplan}
\end{figure}

The roster is defined in Figure~\ref{figure:operationplan} and
displays the requirements of the shift plan. It can be easily seen
that the shifts required for operation remain the same week by week,
with one notable exception: On every third Saturday of the shift
schedule, a different setting is required due to maintenance
work. Because of this, the cycle of the operation plan is set to three
weeks.

The optimization methods applied in the StarFLIP++ environment are all
dependent on an initial solution. It does not really matter whether
the quality of the solution is good or bad, as several studies have
already shown (e.g., in~\cite{Dorn96}), and our experiments
empirically confirmed these results. An initial suboptimal template
problem instantiation ($=$ initial solution) is given. The initial
solution has to satisfy the `hard' requirements of operating hours and
average working hours per week. The repair steps that will be
explained in Section~\ref{section:shiftcons} change the solutions only
in such a way as not to violate these hard constraints.  The
generation of an initial problem instantiation is actually a nice
example of a combinatorial problem in itself. The problem is to find
an initial solution that satisfies the hard constraints of operating
hours and average working hours per week. This problem however is not
considered in the present paper. We now turn to the major parts of the
StarFLIP++ project used to model and solve the presented shift
scheduling problem.


\section{Solving combinatorial optimization problems with
StarFLIP++}\label{section:starflip} 

The following section gives an overview of the StarFLIP++ project. It
puts the system and application presented in this paper into a wider
context. After shortly touching upon the entire StarFLIP++ project, we
concentrate on the part most relevant for the shift scheduling
application.

StarFLIP++ is a library~\cite{StarFLIP97,Slany97b} for real world
decision making. It is a tool for optimization under vague constraints
of different importance using uncertain data. Through the use of fuzzy
computations, compromises between antagonistic criteria can be
modeled. Typical application areas include scheduling, design,
configuration, planning, and classification.  

A production scheduling problem in a steel production plant has been
the key application area for the major part of the development time of
StarFLIP++. Nevertheless, the design of the library has never been
explicitly biased towards a certain application problem. Due to this
fact, an open system evolved that can treat a large variety of
problems with shift scheduling being just one of them.

StarFLIP++ (pronounce: StarFlipPlusPlus; this refers to the fact that
StarFLIP++ stands as a regular expression for all names of the
individual sub-libraries, and all of the latter are based on FLIP++)
was created to investigate real world combinatorial optimization
problems such as the shift scheduling problem described in the
previous section. It was designed and implemented as a family of {\em
C++\/} libraries. StarFLIP++ is composed of the following layered
sub-libraries:

\begin{itemize}
\item FLIP++: the basic fuzzy logic inference processor library.
\item ConFLIP++: the static fuzzy constraint library which recently
has been merged with DynaFLIP++.
\item DynaFLIP++: the dynamic fuzzy constraint generation and
interpreter library for the constraint script interperation (CSI) 
language. 
\item DomFLIP++: the domain knowledge representation library.
\item OptiFLIP++: the heuristic optimizing library; several repair
based heuristics have so far been implemented and tested.
\item CheckFLIP++: the knowledge-change consistency checker library 
that also allows fine-tuning of the configuration parameters of a 
problem.
\item InterFLIP++: the graphical user interface for all other 
libraries, with platform support for X-Windows (XView/OpenLook, Motif)
and MS Windows 3.1/95/NT. 
\item ControlFLIP++: the control center where the interplay between
the other parts is coordinated (mainly data I/O and calling
functionality). 
\item DocuFLIP++: the online documentation available separately for
end-users, knowledge engineers, and programmers, and accessible via
the
World-Wide-Web\footnote{http://www.dbai.tuwien.ac.at/proj/StarFLIP/}
as HTML documents.
\end{itemize}

Furthermore, the following parts are under development:

\begin{itemize}
\item ReaFLIP++: the reactive optimizer as an extension of DomFLIP++.
\item NeuroFLIP++: the neural network extension that allows automatic
tuning of fuzzy membership functions.
\item TestFLIP++: the version control and test environment for the 
complete library set.
\item SimFLIP++: the simulation toolkit library.
\item JavaFLIP++: a major reuse/redesign of the existing StarFLIP++
libraries currently under way in the {\em JAVA\/} programming language.
\end{itemize}

\begin{figure}[th]
\epsfbox{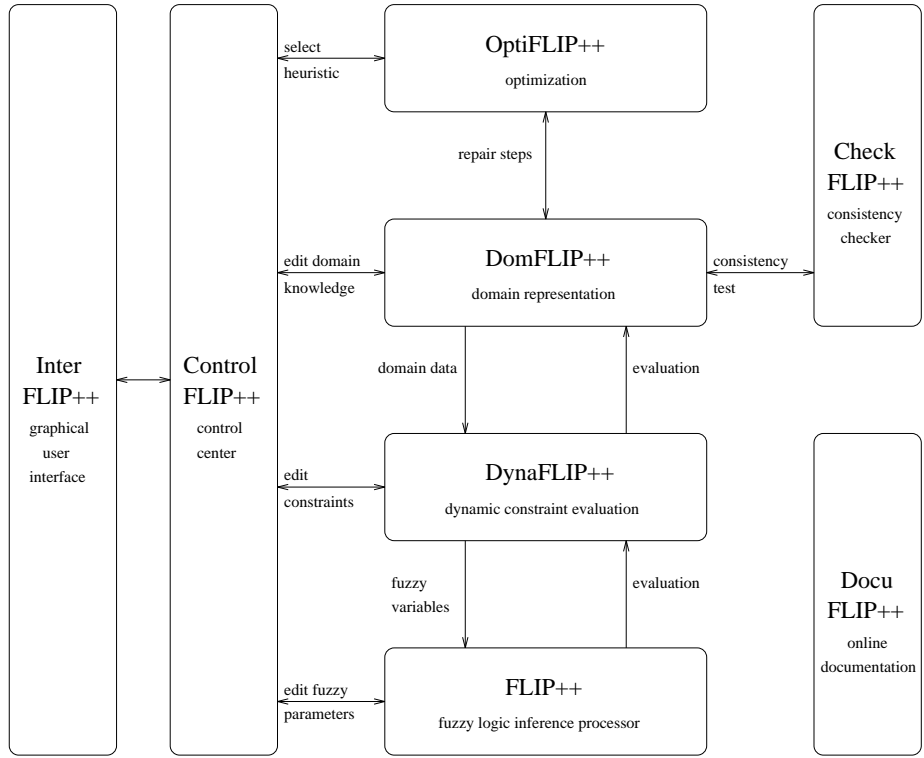}
\caption{Overview on the structure of StarFLIP++.}
\label{figure:starflip}
\end{figure}

Figure~\ref{figure:starflip} shows a view of the layered structure of
StarFLIP++.

These libraries come without domain knowledge base. Therefore, during
a first knowledge acquisition phase, the knowledge engineer describes
the items (e.g., products, workforce groups, \ldots{}) and the logical
objects (e.g., machinery, shift plan tasks, \ldots{}) with their
respective attributes for the environment. This information is stored
using DomFLIP++.  In a second step, the functional relations among
process variables have to be defined. A variety of mathematical
description methods have been implemented for this goal. In a further
step, constraints for these process variables can be entered to define
restrictions in the value domains of these variables. This is done in
two stages: First, static template constraints are defined using the
ConFLIP++ part of DynaFLIP++, for instance to specify a due date
constraint, i.e., the constraint that a generic job will have to be
finished by some time yet to be specified, with a certain gradual
satisfaction defined trough fuzzy variables, terms and associated
membership functions.  Second, rules governing the application and
specialization of such template constraints to particular instances of
the combinatorial optimization problem at hand are defined by the
knowledge engineer in DynaFLIP++. This specialization occurs normally
during optimization time as constraints need to be interpreted to
allow their flexible adaptation to, for instance, a particular number
of jobs that cannot be foreseen at specification time. This is similar
to the use of aggregation functions in spreadsheets or databases.

When this knowledge modeling step is finished, a given instantiation
of a schedule can be constructed from actual process data and, after a
schedule has been proposed, evaluated. After the evaluation of all
constraints, changes on the schedule are usually done in order to find
a more satisfying instantiation. What these changes are and how they
look like is specified by the knowledge engineer in relation to the
optimization methods supported by the OptiFLIP++ library. For
instance, the genetic optimization algorithm uses special genetic
operators such as the crossover operator to perform changes on the
schedule, which are useless in the tabu search type optimization. So
for each optimization algorithm the knowledge engineer wants to apply
to the problem at hand, it is possible to specify a range of
corresponding repair operators that change some parts of the schedule.

Several repair based algorithms were integrated in OptiFLIP++, namely
\begin{itemize}
\item a tabu list min-conflicts repair based hill climbing heuristic,
\item a min-conflicts repair based iterative deepening heuristic, 
\item a min-conflicts repair based random search hill climbing
heuristic, and
\item a min-conflicts repair based genetic algorithm heuristic.
\end{itemize}
All repair based algorithms have several variants and many
parameters. A conflict identification function is used together with a
domain dependent repair operator library to quickly choose the repair
operator that will most probably minimize conflicts for a given
situation. However, the algorithms are independent of this library since
the guidance provided through the conflict identification function is in
all cases combined with a fall-back random strategy if nothing else helps
to find better instantiations.

The following sections illustrate the main modules necessary to define
a new problem instance such as the shift scheduling application. 

\subsection{Modeling fuzzy constraints with ConFLIP++}

\begin{figure}[th]
\epsfbox{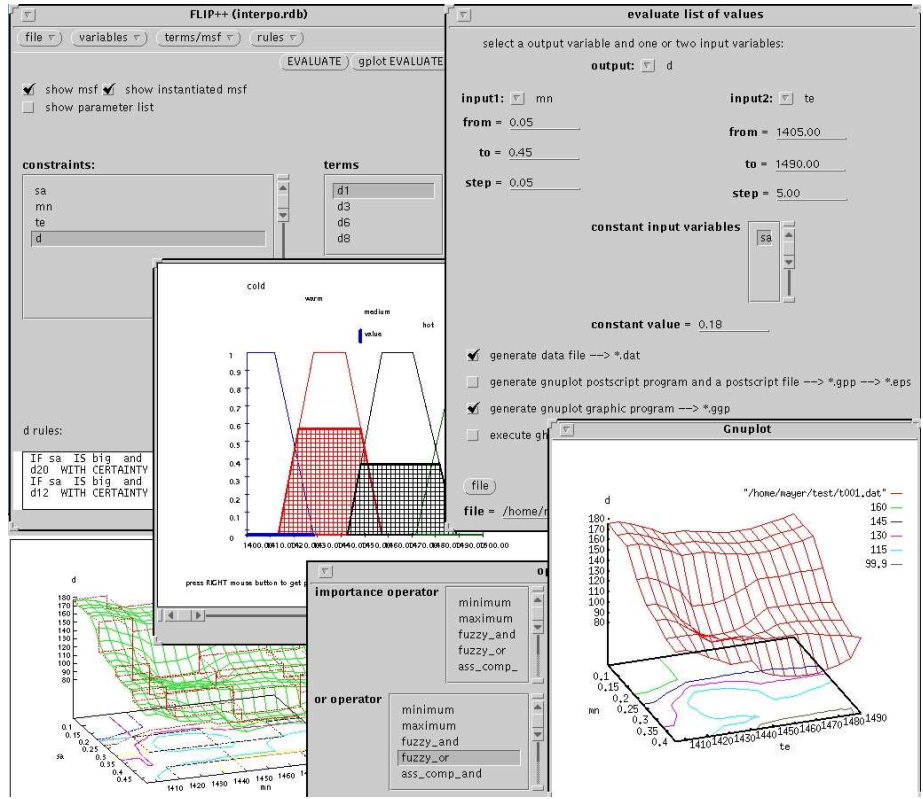}
\caption{Typical screen-shot during an XView interaction with
InterFLIP++.}
\label{figure:uiflip}
\end{figure}

The reusable {\em C++\/} object library ConFLIP++ is a constraint handling
extension to FLIP++, which itself is a general purpose fuzzy logic
inference library. ConFLIP++ was merged into DynaFLIP++ for efficiency
reasons but provides an independent interface with its own
functionality. Because it constitutes the basis of the rest of the
project, we will explain it in this section. First, however, let us
describe it in the context of FLIP++ which handles everything
concerning fuzzification, membership functions, and linguistic
variables. The user can choose between several different fuzzy
inference methods, various priority schemes, different aggregation
operators, and several defuzzification methods. FLIP++ also permits
the graphical editing of membership functions and the easy
manipulation of rule sets. Bonner et al.~\cite{Bonner95} describe for
instance how to solve a fuzzy control problem using FLIP++ alone. The
InterFLIP++ userinterface tool supports all functionality provided by
ConFLIP++ and FLIP++. This includes creating, interactively editing,
saving and reloading named sets of constraints including all
parameters, and evaluating constraints. Figure~\ref{figure:uiflip}
shows a typical screen-shot during an XView interaction with
InterFLIP++. ConFLIP++ thus serves as a knowledge engineering tool in
which domain knowledge can be stored, manipulated, and used for
reasoning independently from the rest of the program.

In ConFLIP++, the first step is creating simple constraints such as
the following taken from the steel making application:
\begin{center}alu-cntnt $\leq$ 0.08
\end{center}
and naming them in the case of the example for instance `alu-cons'
using the objects and methods defined in ConFLIP++. The aim is to
catch vagueness in constraint-equations where the $\leq$ sign is not
meant to be interpreted in its strict mathematical sense, but such
that `smaller' violations are acceptable. What these `smaller'
violations could be has to be defined explicitly (and precisely)
through the membership functions associated to the `terms' of the
variable as defined below. Additionally, ConFLIP++ is able to handle
uncertainty about the exact value of `alu-cntnt', which is possible by
propagating possibility distributions instead of defuzzified
values. The operators to infer values and to aggregate several
constraints are then applied to fuzzy values, which can always be
represented as membership functions. This capability to model {\em
with accuracy\/} vague relations and uncertain data is the major
contribution of fuzzy and possibilistic logics.

In our example of a simple constraint, the aluminum content
`alu-cntnt' is a so called {\em linguistic variable}, a generalization
of the conventional concept of a variable. A linguistic variable has a
finite set of terms, which are mapped to an interval of real numbers by a
membership function. By a linguistic variable we mean a variable whose
values are words or sentences in a natural or artificial language
rather than numbers. In ConFLIP++ fuzzy sets like linguistic terms are
represented by the {\em ParameterSet-object}. For example the fuzzy set
{\em temperature\/} has the linguistic terms {\em cold, medium, warm},
which are mapped to intervals of real temperature values by the appropriate
membership function. The object to model a simple constraint has the
following structure:
\begin{center}\small
{\em ConstraintCompare(name, importance, dilatation, comment,
variable, compare\_operator, compare\_value)}
\end{center}

In the next step, several such constraints are logically combined, i.e.,
they are aggregated by one of the aggregation operators, such as AND
and OR, which for example could be implemented as {\em minimum\/} and
{\em maximum\/} operators, to build more and more complex constraints
with a hierarchical structure.  The object to model such complex
constraints has the following structure:
\begin{center}\small
{\em ConstraintConcat(name, importance, dilatation, comment, 
constraint, constraint, concat\_operator)}
\end{center}
where dilatation is the type of the constraint (either crisp, fuzzy,
or mixed), and constraint is either a ConstraintCompare or a
ConstraintConcat. ConFLIP++ then automatically creates a rule-set out
of default or user-defined terms of sets for standard linguistic
variables, standard rule set tables, standard membership functions for
the term sets, default priority values, and various default operators
using FLIP++. FLIP++ is repeatedly called later to evaluate the
constraints for some instantiations of the free linguistic variables
appearing in the constraint. Additionally, the system checks the
scores of all constraints having a priority different from zero as
well as of their constituent sub-constraints before these constraints
are aggregated to find out whether a hard constraint violation
occurred (evaluation score equals zero) in order to invalidate
instantiations that crossed the hard barrier of the corresponding
constraint, which is not allowed.

The rule-set is built for instance such that, if the first linguistic
variable is compared to its term `positive\_big', and the involved
inequality is `variable $\geq$ constant', and another linguistic
variable is compared to `zero', and the constraints corresponding to
the two linguistic variables are concatenated by `or', then the
resulting term for the aggregated rule is `very\_good'. The latter
term comes from the predefined template term set \{`very\_good',
`good', `zero', `bad', `very\_bad'\}.

The human expert will usually have to fine-tune the automatically
created rule-set and the membership functions associated to the
terms. However, it is possible to store user defined standard sets of
term sets and the associated set of semantic rules. Additionally,
fuzzy methods are quite robust, such that the exact determination of
the membership functions is not essential. The predefined triangular
membership functions often perform well in a first
approximation. Nevertheless, one reason that makes fine-tuning
necessary is that ConFLIP++ has no a priori domain knowledge. If the
constraint is `alu-cntnt $\leq$ 0.08', some generated default
rules are for instance:
\begin{quote}
IF alu-cntnt is positive\_small THEN alu-cons is zero\\
IF alu-cntnt is positive\_medium THEN alu-cons is bad\\
IF alu-cntnt is positive\_big THEN alu-cons is very\_bad\\
\end{quote}
The really important object of ConFLIP++ is {\em
SetOfConstraints}. Each SetOfConstraints has a name and a list of
constraints. Furthermore, it needs a rule-set-object, tables for
concatenating and comparing constraints with the appropriate operators
defined in the {\em OperatorSet}, and a parameter set describing fuzzy
linguistic variables.

The next step is the evaluation of a constraint. The evaluation
happens according to the rule-set of the constraint. First, the free
linguistic variables have to be given values, the latter being either
defuzzified real numbers or possibility distributions.  The
evaluation function returns by default a defuzzified value that
describes the degree of satisfaction of the constraint with the given
values. The ConFLIP++ object that holds evaluated constraints for
further operations is the {\em SetOfEvalConstraints}.

The human expert can influence the decision making behavior of
ConFLIP++ in various ways. After a constraint knowledge base has been
compiled, it can be copied and the copy can be edited. First, the
human expert can select one of several aggregation, implication, and
defuzzification operators. The weighing scheme can be chosen as
well. Of course, the individual membership functions and priorities of
the constraints can be graphically edited. For instance, it is easy to
selectively edit the constraint responsible for the observation of
due dates. These changes will immediately take effect on the decision
function. To ease configuration of a complete constraint knowledge base
built up from scratch, the default values for all these parameters are
pre-specified in a way that seems to apply reasonably well to most
cases. However, the human expert can later soften or harden all those
constraints that have not yet been fine-tuned on an individual base. In
such a case, ConFLIP++ searches the complete knowledge base for
membership functions that make the decision making behavior of the
constraint knowledge base fuzzier or crisper.

\subsection{Evaluating dynamic constraints with DynaFLIP++}

Dechter and Dechter~\cite{Dechter88} introduced the term {\em dynamic
constraint network\/} to deal with changes from one constraint network
to another one, such that new facts about the environment can be
modeled. While this issue is taken care of in the CheckFLIP++ part of
StarFLIP++ (see Section~\ref{section:checkflip}), in DynaFLIP++ we
want to focus on a different problem, namely on the {\em
dynamic\/} generation of constraints at runtime, based on environment
data and so called template or static constraints. Thus, our use of
the term {\em dynamic constraint\/} is unrelated to the dynamic
constraint networks defined in~\cite{Dechter88}.

The DynaFLIP++ library subsumes ConFLIP++, the latter being needed to
formulate static fuzzy constraints and then operate on them.
DynaFLIP++ wraps additional functionality around these fuzzy
constraints and reaches this functionality up to DomFLIP++ where the
actual domain knowledge is processed and optimization is performed.
This additional functionality has become necessary because of factors
encountered in many real world combinatorial optimization problems, of
which we only became aware of after trying to handle several different
problems. Since special consideration is given to the reusability
aspect of all libraries, we found that static fuzzy constraints are
good to model the aspects associated with partial satisfaction,
compromising, and relative importance of constraints.  However, other
aspects require dynamic generation of constraints from template
constraints. For instance, the actual number and kind of constraints
often depend upon the current instantiation of the problem that is to
be optimized. With each repair step in the optimization process, the
structure of this instantiation may change. Therefore, the structure
of the constraint evaluation tree has also to change.  For example, it
is possible that a job with an associated delivery date is exchanged
with another job that has no delivery date, therefore this constraint
must not be evaluated for the second schedule.

Another aspect is that it is normally not useful to tune each
constraint separately.  Instead, a static constraint is tuned for a
selected reference value, and DynaFLIP++ then uses this static
constraint to generate a dynamic constraint adapted to the actual
situation. Again, looking at the delivery date example, this implies
that the template is a static constraint that is tuned around the
value zero, with an appropriate fuzzy distribution around it.
DynaFLIP++ then specializes this to an actual time in the scheduling
horizon.  This dynamic adaptation is mostly harmless for simple
constraints such as delivery dates, but becomes more tricky when
complex constraints are involved.  A more complex example, also taken
from the steel making domain, would be the duration of tundish life
expectancy.  The tundish, a part of the caster, has to be maintained
after approximately 240 minutes, but this length can vary between 100
and 300 minutes.  The problem is that the attributes of a finite
number of jobs must be aggregated, in this case by adding their
durations, without knowing at the time when the static constraint is
defined how many jobs will have to be eventually aggregated. Their
number can only be determined dynamically at optimization time by
looking up compound data values in the actual schedule
instantiation. These aggregation operators are similar to those found
in spreadsheet software that process a range of values. At this point
we would like to clarify the meaning of the term {\em aggregate\/}
that in the context of steel making it is a synonym for large
metallurgical equipment such as continuous casters or blast
furnaces. On the other hand, {\em aggregation\/} stands for the
conjunction of constraints by soft AND and OR operators as we have
seen before.  Additionally, {\em aggregation\/} as in {\em aggregation
operators\/} in the context of this section can also be found in the
field of constraint databases as described by Kuper~\cite{Kuper93}.
This is because the number of arguments in constraints evaluated by
DynaFLIP++ depends on conditions that can be checked only at execution
time.

A difficulty arises because DynaFLIP++ contains absolutely no
knowledge about the domain. Using DomFLIP++, optimization structures
are defined together with their associated constraints, using the
knowledge engineering features of DynaFLIP++. Additionally, relations
between attributes of items in these optimization structures and the
static constraints must be defined using a special rule based language
that is interpreted at evaluation time by DynaFLIP++. A rule,
slightly simplified but taken from the actual steel making domain,
illustrates this mechanism as given in Figure~\ref{figure:cccx}.
\begin{figure}[th]
\begin{center}
\begin{minipage}{8cm}
CONDITION:\\
\hspace*{3mm} $\langle$Job$_i$,Job$_{i+1}$$\rangle\in$ CC$_3 \,\wedge$\\
\hspace*{3mm} quality separation $\not\in\langle$Job$_i$,Job$_{i+1}$$\rangle \,\wedge$\\
\hspace*{3mm} tundish change $\not\in\langle$Job$_i$,Job$_{i+1}$$\rangle \,\wedge$\\
\hspace*{3mm} CC setup $\not\in\langle$Job$_i$,Job$_{i+1}$$\rangle$\\
CONSTRAINT:\\
\hspace*{3mm} \mbox{Chemical\_Compatibility\_CC$_3\langle$Job$_i$,Job$_{i+1}$$\rangle$}
\end{minipage}
\end{center}
\caption{A rule specifying when and how a template constraint is used.}\label{figure:cccx}
\end{figure}
Here,
`Chemical\_Compatibility\_CC$_3\langle$Job$_i$,Job$_{i+1}$$\rangle$'
stands as a macro that links several attribute values of Job$_i$ and
Job$_{i+1}$ to linguistic variables defined in the pre-tuned static
constraints. For the actual implementation, the whole definition of
this macro must be specified.  A sketch of the definition, where only
the formulas involving the chemical element carbon are detailed, is
presented in Figure~\ref{figure:ccc}, 

\begin{figure}[th]
\hrule
\vspace*{2mm}
Job$_i$.C.high $\stackrel{\rm soft}{<}$ alloy\_limit.C
$\quad\stackrel{\rm soft}{\wedge}\quad$ Job$_{i+1}$.C.high
$\stackrel{\rm soft}{<}$ alloy\_limit.C $\quad\stackrel{\rm soft}{\vee}$\\\smallskip
Job$_{i+1}$.C.high $\leq$ Job$_i$.C.high $\quad\stackrel{\rm
soft}{\wedge}\quad$ Job$_i$.C.low $\leq$ Job$_{i+1}$.C.low $\quad\stackrel{\rm soft}{\vee}$\\\smallskip
Job$_i$.C.high $\leq$ Job$_{i+1}$.C.high $\quad\stackrel{\rm
soft}{\wedge}\quad$ Job$_{i+1}$.C.low $\leq$ Job$_i$.C.low $\quad\stackrel{\rm soft}{\vee}$\\\smallskip
overlapping.C$\langle$Job$_i$,Job$_{i+1}\rangle$ $\stackrel{\rm
soft}{\geq}$ pair\_limit.C\\[3mm]
\hspace*{3mm} {\large (}where\quad overlapping.C$\langle$Job$_i$,Job$_{i+1}\rangle:=\\\hspace*{21mm}\max(0,\min($Job$_i$.C.high,%
Job$_{i+1}$.C.high$)-\\\hspace*{33mm}\max($Job$_i$.C.low,Job$_{i+1}$.C.low)){\large
)}\\[3mm]
\hspace*{3cm}$\LARGE\vdots$ \qquad similar for other chemical elements.
\vspace*{2mm}
\hrule
\caption[]{Sketch of macro definition for a restriction to carbon of Chemical\_Compatibility\_CC$_3\langle$Job$_i$,Job$_{i+1}\rangle$.}\label{figure:ccc}
\end{figure}
thus giving an idea of what kind of
dynamic adaptations must be computed.  The actual compatibility
encompasses 12 more chemical alloying elements, the degassing
procedure in the secondary metallurgy aggregates, as well as the
casting format between the jobs.

To evaluate a given instantiation of a partial constraint satisfaction
problem, a decision function aggregating all the constraints with
their respective priorities, using an appropriate aggregation operator
and a corresponding weighing scheme, must be established.  Whereas
the representation of template constraints is handled with the
ConFLIP++ library, we present in this section the DynaFLIP++ library
responsible for efficiently establishing a new global constraint
representation for a specific instantiation of the problem.  This
global constraint will result in a highly structured constraint tree
for the whole schedule. The constraint evaluation function will return
the weighted global satisfaction score based on the current schedule
and the constraint evaluation tree. The nodes of this dynamically
constructed tree nodes are weighted aggregation operators (in the
simplest case conjunctions) and the leafs are ConFLIP++ objects
representing individually fine-tuned static constraints, again taken
from the steel making domain.  DynaFLIP++ is able to use most of the
framework provided by ConFLIP++ to efficiently compute the evaluation
scores for a new schedule.

When optimizing, it is often advisable to introduce an additional
measure into the decision function dependent upon whether the current
instance of the combinatorial optimization problem contains certain
difficult items. In the scheduling context, this would mean that if
the scheduling of these jobs is not introduced as a bonus into the
decision function, these jobs might never be considered for actual
scheduling. There usually exists a non empty pool of waiting jobs, and
only a subset of jobs from the pool can be scheduled immediately.
Therefore, the danger is that some difficult jobs will remain in the
pool forever unless additional measures are taken. It is clear that
this `difficulty' or `importance' of a job must increase over the time
for which it is still reasonable to `produce' it, to allow its
eventual scheduling.  The easiest way to introduce this `difficulty'
is to formulate a corresponding constraint with an associated priority
that will represent these difficult jobs and which will therefore be
represented by another branch of a certain constraint type. Thus, the
`difficulty' of jobs will be one criteria considered when the partial
constraint satisfaction problem is optimized.  The same applies
equally to other partial constraint satisfaction problems such as
those encountered in design or planning.

 To build up the evaluation tree, DynaFLIP++ has to consider the
domain structure with its aggregates and scheduling objects, so that
the evaluating tree is built up analogous to the hierarchical
structure of the modeled application. The structure of the scheduling
objects depends on the application they are designed for. Considering
production scheduling in industrial environments as a special
combinatorial optimization problem, we encounter different types of
imprecision, stemming from constraints that are blurred in definition
and include vagueness and uncertainty. We can imagine that an
operation on the schedule may start a `little' earlier and that
`small' deviations of optimal values may be acceptable. The scheduling
object `Order' then has attributes such as {\em plant, plant-mark,
throughput, weight, format, thickness, speed, slab-group,
chemical\_elements, delivery\_date, \ldots{}} and associated
constraints like

\begin{center} domain constraint: out\_date $\leq$ delivery\_date
\end{center}

In this case, an aggregate could be the {\em continuous caster CC-4},
with constraints concerning {\em tundish durations, average
throughput, or setup-restrictions}.  Of course there would exist a
variety of other constraints, such as compatibility constraints,
capacity constraints, or temporal constraints. To evaluate a schedule
for one aggregate, DynaFLIP++ has to first create and instantiate all
SetOfConstraints for the domain objects which were specified by the
knowledge engineer. In a second step, all the variables that have
restrictions in form of constraints have to be computed. This is done
by evaluating the computation clauses designed parallel to the related
constraints. Of course all variables have to be computed before they
are compared to constraints. The next step is the adaptation of
template constraints to the actual situation on the schedule. We can
imagine that a template compare value has a fuzzy distribution around
zero, but the real compare value may be an aggregation of other
processes variables and would have another value.

\begin{center} template constraint: fuzzy\_var\_foo1 $\leq$ 0\\
specialized constraint: out\_date - delivery\_date $\leq$ 0
\end{center}

On the evaluation of each constraint, DynaFLIP++ invokes the
evaluation mechanisms of ConFLIP++. The evaluated constraints are put
into the SetOfEvalConstraints, where they can be aggregated as
described in Figure~\ref{figure:cccx}. When all relevant constraints
determined by the rule based language are evaluated, the overall
evaluation score is returned to DomFLIP++. Further, we are interested
in the biggest violations on the schedule, so we select bad
evaluations at runtime and put them into an appropriate data
structure, which is later used by DomFLIP++ to make changes on the
schedule in order to avoid these violations. The repair steps depend
on the optimization algorithm used by DomFLIP++. If a local change on
the schedule has occurred, DynaFLIP++ has to check where the changes
took place, in order not to build up a complete new evaluation tree,
but only to recompute those parts of the schedule that have been
changed.  This reuse of already computed data structures will, similar
to a caching mechanism, influence the runtime behavior of the
evaluation process.

The most important object of DynaFLIP++ is the evaluation tree which
contains the specialized constraints, the evaluated constraints, and
the aggregation of the latter. The variables with their actual values
and the violations are stored in a separate structure.

\subsection{OptiFLIP++ and ControlFLIP++}

To guide the search of the OptiFLIP++ algorithms as discussed
in~\cite{Slany96a}, it is necessary to identify the constraint with
the worst weighted evaluation, i.e., the severest conflict which can
be attacked to minimize conflicts. This can be considered as a side
product of evaluating the current instantiation. It corresponds to
computing the evaluation using the minimum operator, and more
importantly, to remember the constraint involved in the minimal
weighted evaluation. This constraint represents the largest conflict
for the current instantiation.  Often the constraint corresponds to a
general feature of the instantiation and cannot be attributed to a
specific part of the instantiation. Depending on the repair operators
available to the repair based constraint satisfaction algorithms, it
can be helpful to find additionally the second largest and third
largest conflict. Generally, the search should return the largest
conflict being of a type that can be handled by an available repair
operator. When DynaFLIP++ has to generate a new dynamic constraint
representation for a given instantiation, it computes the individual
`leaf' constraints by calling ConFLIP++ repeatedly with new
variable instantiations on one of the stored reference constraints,
and stores the results in an intermediate form that can be used by
ConFLIP++ for further aggregation. This ensures a relatively efficient
processing of the constraints since the sometimes very large data
structure of a static constraint can be reused for all dynamic
constraints of its type. At the same time, DynaFLIP++ sorts all the
computed intermediate evaluation scores, together with type
information, for later selection of `good' repair operators.

We designed our constraint satisfaction engine StarFLIP++ with a
real world problem in mind, namely the scheduling of fine grained
production in a steel making plant. Typically, this involves more than
a thousand binary soft and hard constraints and a matching number of
variables with continuous domains. The number of constraint checks
until a satisfactory solution is found ranges in the several hundred
thousands. In view of these numbers, it is clear that complexity
issues play at least as important a role as the one played by
good knowledge acquisition tools. As mentioned above, we therefore
adopted {\em repair based\/} algorithms that have been shown to be
very efficient strategies for large constraint satisfaction
problems. Minton et al.\ \cite{Minton92} could find solutions in less
than four minutes on a Sparc workstation 1 for the {\em million\/}
queens problem, while the best general backtracking approach (found in
an empirical study by Stone and Stone \cite{Stone87} to be a
most-constrained backtracking algorithm) became intractable for $n >
1000$.  Minton et al.\ \cite{Minton92} even found that their repair
based method exhibits linear time and space complexity for large
$n$. The {\em min-conflicts heuristic\/} combined with a {\em repair
based hill climbing heuristic\/} specifies that, starting from an
initial suboptimal solution, the system attempts to minimize the
number of constraint violations after each repair step. Minton et al.\
\cite{Minton92} showed convincingly that for certain problems, the use
of the additional knowledge gained from operating on complete but
suboptimal solutions instead of building solutions from scratch as in
constructive approaches pays off well.  Such repair based heuristics
perform orders of magnitude better than traditional backtracking
techniques.  Though repair based methods can be combined with many
general search strategies, they found that hill climbing methods were
especially well suited for the problems they investigated.  While this
result is very nice for a {\em general\/} constraint satisfaction
technique, the apparently not well known fact that Abramson and Yung
\cite{Abramson84} found a constructive method to solve the general $n$
queens problem with {\em linear time\/} complexity should not be left
untold. Though this implies that $n$ queens is not an intractable
problem, it shows that general repair based algorithms often attain
almost the optimal theoretical complexity, which seems to be untrue
for general constructive backtracking algorithms.  Statistics and a
detailed analysis of the different algorithms can be found in
\cite{Slany96a}. All repair based heuristics were much faster and
yielded better results than the constructive approach that was
evaluated using the same configuration parameters on real world
instances of combinatorial optimization problems.

Another point speaking in favor of repair based approaches for
combinatorial optimization problems is that these algorithms do not
need to prune away search branches and can still be very
efficient. While pruning as described in \cite{Freuder92} is well
suited to solve classic constraint satisfaction problems, its
application to combinatorial optimization problems poses several
problems. For one, compromises can only be evaluated by looking at all
constraints.  Additionally, real world problems actually often cannot
be completely described by constraints, because for instance future
events cannot be predicted in scheduling.  Therefore, it is sometimes
desirable to reject optimal solutions in favor of slightly worse but
{\em robust\/} solutions, in the sense that small alterations in the
actual execution of, e.g., a schedule, still belong to good
instantiations, while the superficially best solution is surrounded by
very bad ones. By early pruning, it is of course impossible to
investigate such a situation.

\subsection{Defining an optimization problem with DomFLIP++}

The DynaFLIP++ library was located between ConFLIP++ and the domain
knowledge representation library DomFLIP++, which is a description
tool for the environment that has to be optimized. The structure of a
domain holds a list of aggregates, which themselves hold a schedule of
objects with their respective attributes and variables.  Additionally,
on each level one or more SetOfConstraints can be specified in order
to describe relations and restrictions on the process
variables. DomFLIP++ is also responsible for repair steps on a badly
evaluated schedule, using a list of violations and badly evaluated
variables, which are computed at runtime by DynaFLIP++, and the
optimization algorithms supported by OptiFLIP++.

DomFLIP++ is the knowledge representation module of the StarFLIP++
project. StarFLIP++ focuses on optimizing combinatorial problems that
can be expressed as multiple criteria problems. It uses fuzzy
constraints to model optimizing criteria and applies various iterative
improvement techniques such as Tabu search, genetic algorithms, and
iterative deepening to the problems. It allows the definition of new
optimization problems by aiding the domain engineer in the design of
the structure of a new problem at hand. Generally, a division between
domain dependent and domain independent methods and data structures
characterizes the structure of DomFLIP++. While the domain dependent
data structures are specific to the problem, the domain independent
part is provided as a framework by the library. Moreover, there is a
domain independent interface to other StarFLIP++ modules such as
OptiFLIP++, DynaFLIP++, and CheckFLIP++. After each iteration in the
optimization process, the considered instantiations of the problem are
evaluated. Each evaluation produces a list of evaluated constraints
and hence provides hints on violations of requirements. For each
constraint, modification operators, also called repair steps, are
defined that can be used to increase the score of the constraint in
further iterations of the optimization. A domain can be fine-tuned
through modifying of constraints and their fuzzy representation,
changing the choice of repair steps, and varying optimizing
parameters. A well tuned domain can then be successfully
optimized. The shift scheduling domain presented in this paper is a
fruitful area of investigating the power of DomFLIP++. This is due to
its variety of constraints, inherent fuzziness of requirements, and
large search space that recommends the application of heuristics.

\subsection{Changing domain descriptions using
CheckFLIP++}\label{section:checkflip} 

Slany \cite{Slany96a} has shown that the ordering behavior of the
priority values can be chaotic, in the sense that {\em small\/}
changes in the knowledge base can have {\em large\/} effects on the
ranking of solutions. While this seems rather counterintuitive at
first, it makes sense after looking closer at the situation. One
example in \cite{Slany96a} has the ranking of several partial
solutions inverted because of unforeseeable interactions between
operator fix-points and weights of constraints, i.e., the rankings of
instantiations (large results) are sensitive to certain threshold
values in the knowledge base (small changes), so changes can produce
unpredictable, chaotic results. In light of the link between
non-monotonic logics and combinatorial optimization problems as
developed by Brewka et al.\ \cite{Brewka92}, such non-monotonic
reactions to changes in knowledge bases seem to be an obvious result.

Since there is no unique way to compute weights of constraints for a
given problem, there is also no clear way to relate weights to the
wishes of the field expert regarding priorities, other than
experimenting and fine-tuning by testing different variants. It is
possible to completely change the ranking behavior of weights by
switching to another aggregation operator or to a different weighing
scheme. The examples given in \cite{Slany96a} demonstrate that
fine-tuning of the parameters for a combinatorial optimization problem
is absolutely necessary in order to obtain meaningful
results. Section~\ref{section:checkflip} indicates how this tuning can
be done rationally while avoiding inconsistencies with former
decisions. Since the method is based on trial-and-error, and since
test cases are used to implicitly limit changes in a knowledge base,
the method effectively helps to harness the chaotic behavior described
above.

A major concern in decision making problems is how to correctly elicit
knowledge from human experts. The project comprises a method of
eliciting the criteria's importances from human experts. Especially
when many human experts have to agree on a problem description such as
the rules involved, the importances of certain criteria, etc., it is
important to have a method that allows to make reasonable and
consistent changes to the parameters of the problem description. The
test implemented in the CheckFLIP++ part of the project highlights all
inconsistencies in configuration changes. The test also helps to
evaluate the sensitivity to configuration changes and provides a
possible way to allow automatic learning of problem descriptions.

Freuder and Wallace \cite{Freuder92} observe that weakening
constraints in effect means creating a different problem. In the
present section, we have shown that it is often unclear which problem
we should solve, and that small changes in parameters describing a
combinatorial optimization problem might cause large and unforeseeable
changes in the corresponding solutions. Therefore, it seems justified
to ask what kind of changes should be allowed and what implications
these changes might entail.

An answer to the problem of making sure that fine-tuning is done
consistently with earlier decisions is to adopt a consistency test for
configuration changes. Such configuration changes could be changes in
the priorities between constraints, adopting a new aggregation
operator, changing hard barriers, changing membership functions, or
changing the logical structure of constraints. Basically, this change
together with the test produces a new ranking for a given set of new
instantiations, while observing predefined rankings for a set of old
reference ranking of pairs of instantiations.  The mechanism works
such that, if the human expert is dissatisfied with a ranking produced
by the system, he or she can slightly change the weights of some
constraints, or the exact form of some membership function (e.g., to
specify that a hard barrier is actually located slightly higher), or
any other parameter of the problem, such as the aggregation operator
used. A consistency test will then check whether the new configuration
is consistent with the rankings for a set of reference pairs of
instantiations.  This is done by applying the new configuration, e.g.,
the set of new weights, to all the old ordered pairs of
instantiations, and by calculating their evaluation scores with this
new configuration. If for each reference pair the order between the
two reference instantiations remains unchanged, this indicates that
the new configuration does not invalidate any previous reference
ordering. It is compatible with all decisions made in the past that
became reference ranking pairs.

If one reference ranking pair is ranked in the opposite order, this
means that either the new configuration is {\em wrong\/} and has to be
changed again, or that some reference ranking pairs are obsolete and
should therefore be removed from the reference ranking pair database.
In both cases, an inconsistency among the reference rankings and the
new ranking is pointed out. This inconsistency has to be resolved such
that the resulting system makes rational, predictable, understandable
and self consistent decisions. The probability that the inconsistency
is due to noise in the problem description and should therefore be
neglected is zero, since all reference rankings have been generated
with the explicit aim to change the configuration in order to give
them a certain, new order. An inconsistency can point to earlier
errors in configuration changes. Since each change is done under
supervision, usually by a human expert, and changes are normally only
adopted with the explicit goal to produce a different ordering, the
inconsistency cannot be attributed to noise.  Whether such a decision
making behavior can be termed {\em objective\/} or {\em subjective\/}
depends on other factors.  However, it is usually possible to lead
{\em several\/} human experts to agree on a common, undisputed subset
of some reference ranking pairs of instantiations, or at least to
establish several different sets that correspond to configurations
which can be further characterized by and saved for later use under
such names as, e.g., for scheduling combinatorial optimization
problems, `risky/cost-cutting', `highest-quality',
`observe-temporal-constraints', `standard-mix', etc., indicating their
general tendency for decision making. This makes clear that there is
no notion of a {\em best\/} combinatorial optimization problem in our
approach, but that several combinatorial optimization problems
optimizing a solution of a real world problem from slightly different
points of view can coexist. The corresponding last configuration is
saved together with these reference ranking pairs of instantiations as
one knowledge base.  Of course, not all intermediate stages have to be
stored permanently.  This permits modeling the intentions of the human
expert with maximal flexibility while ensuring rational and
predictable behavior after changes in the configuration.

If the new configuration is adopted, the best solution {\em before\/}
making the configuration change and the best solution {\em after\/}
making the configuration change become a new reference ranking pair
added to the new database associated with the new configuration. In
this pair, the best solution {\em after\/} making the configuration
change is ranked first, and the best solution {\em before\/} making
the configuration change is ranked second. All data influencing the
overall decision function must be stored together with the pair to be
able to apply the resulting new decision function in the old context,
given the new configuration. \cite{Slany96a} contains a listing of the
consistency test in procedural form.

Human experts can specify implicitly the overall configuration of the
constraints by asserting a set of `normal' reference rankings. The
easiest way to apply the heuristic that establishes consistent
configuration parameters for the constraints is to let the human expert
do parameter changes, and to later check them out with the introduced
consistency test.

\subsection{Automatic knowledge acquisition}

Huard and Freuder \cite{Huard93} view constraint knowledge base
debugging as a partial constraint satisfaction problem in itself. If
the constraint knowledge base is erroneously over-constrained, a change
that entails a small number of new solutions is more in keeping with
Occam's Razor than one that entails many. However, Huard and Freuder
\cite{Huard93} consider only over-constrained networks, while we are
interested in finding a combinatorial optimization problem model that
approximates as closely as possible the {\em implicit\/} problem at
hand, thus leading us to move from one combinatorial optimization
problem to another instead of moving from an over-constrained
constraint satisfaction problem to an approximating combinatorial
optimization problem. Similar to our approach, Huard and Freuder
\cite{Huard93} work in cooperation with a human expert.  This permits
the human user to interactively play {\em what-if\/} games, i.e.,
allowing the expert to see how decision making behavior evolves as
changes are made to the combinatorial optimization problem model. On
the one hand, Huard and Freuder \cite{Huard93} allow only {\em one\/}
constraint to be weakened, while our approach is able to cope with any
kind of change. On the other hand, our method so far does not make any
suggestions for knowledge change, while the knowledge assistant
proposed by Huard and Freuder \cite{Huard93} does. In general, the
`inverse' problem of finding an appropriate combinatorial optimization
problem model given a certain a priori optimal solution, is extremely
difficult because of the multitude of changes that could actually
occur.  Not without good reason do Huard and Freuder \cite{Huard93}
limit changes to only one weakening of one constraint and apply it to
rather small problems. The difficulty is that humans easily overlook
some constraints, especially when the number of constraints is large
and the constraints are only vaguely defined.  Therefore, the
subjective `better' ranking obtained a priori from a human expert will
often objectively not be better than the instantiation found by the
system because the human expert forgot some constraints, thus forcing
the system to learn suboptimal decision making.  Therefore, the
fine-tuning scenario, where human experts repeatedly change constraint 
parameters such as weights by hand and then compare the respective
best solutions, is much better suited to establishing the best
configuration for the problem. This certainly comes from the fact that
human expert do often have an intuitive notion of `good' and `bad'
solutions of combinatorial optimization problems without being totally
aware why they think so. However, it is an open research problem
whether this fine-tuning can be fully automated when an {\em
objective, not prone to human error, a posteriori meta-evaluation\/}
is used, such as one guided by results of quality evaluations.

According to Freuder and Wallace \cite{Freuder92}, such meta
constraints, as for instance induced by the consistency test proposed
in Section~\ref{section:checkflip}, ``are reminiscent of the concept
hierarchies that provide initial bias in machine learning settings,
and indeed it is intriguing to think of the constraint satisfaction
process as a form of concept learning, synthesizing a relationship
from positive and negative information.''

Future work lies in comparing our work to approaches from knowledge
acquisition (e.g., human expert models, cooperative knowledge base
tuning), machine learning (e.g., case base reasoning), as well as
model based diagnosis (e.g., McIlraith and Reiter \cite{McIlraith92}
study the design of tests whose outcomes confirm or refute a
hypothesis).

Huard and Freuder \cite{Huard93} test their knowledge elicitation
method on random problems. However, they start from an idealized
constraint satisfaction problem $P$ that must be approximated; our
method is useful to find an unknown $P$, therefore random
combinatorial optimization problems do not help. We currently believe
that our method can only be tested through satisfactory application to
real world problems such as the steel making application or the shift
scheduling problem presented in Section~\ref{section:shiftpb}.


\section{Shift planning constraints and repair steps}\label{section:shiftcons}

We now come back to the application of StarFLIP++ concepts as
described in Section~\ref{section:starflip} to the shift scheduling
problem we introduced in Section~\ref{section:shiftpb}.

The constraints of the shift scheduling problem define certain
requirements. Normally, constraints are of
dynamic nature, i.e., their concrete instantiations depend on the
instantiation of the problem. In our application this means that
different shift schedules lead to different constraint instantiations
of the same type of constraint. Hence it is necessary to define
constraints in a language-like style that is based on the use of
variables. Each evaluation run of DynaFLIP++ is based on variables
whose values are fed by the problem instantiation of DomFLIP++.

In the following we will shortly explain typical constraints that have
been implemented and that will serve to illustrate the introduced
concepts.

\begin{itemize}
\item {\bf Constraint of even distribution of working hours}

This constraint tries to guide the evaluation towards smooth shift
schedules, meaning that deviations between weekly working hours should
be relatively small. It computes the difference in working hours
between consecutive weeks and detects those pairs of weeks where the
difference is comparatively high, so that they will be used as
possible starts for a repair step. This should lead to a more evenly
distributed sequence of shifts and hence working hours.

\item {\bf Constraint for weekends}

Weekends are a crucial area with most shift scheduling
problems. Often, a human expert has to pay especially attention to
this area of the shift schedule. One constraint simply checks the
number of free weekends for one subgroup and leads to better
evaluations the more free weekends there are.

\end{itemize}

Repair steps represent the modification operators of a problem. They
allow to move from one valid instantiation of the problem to another
one. Such modifications are the basis for every optimizing algorithm
that uses iterative improvement techniques to arrive at --- in terms of
evaluation --- better problem instantiations.

Generally, it is easy to specify certain types of repair steps for a
problem. In our example, the definition of repair steps is almost
entirely dictated by the plan of operation, i.e., the roster. With the
use of the roster as depicted in Figure~\ref{figure:operationplan}, it is
possible to define for each day what kinds of shift and how many of them
must be allocated. Due to performance reasons, repair steps should be
preferably kept simple since they are heavily used during the
optimization process. Mostly, the modifications for a problem can be
broken down to simple swap or move operations. This has also been
shown for the application in the steel production domain
in~\cite{Hendrysiak96}.

Repair steps represent modifications of the shift schedule. Without any
guidance these modifications would take place randomly. To avoid that,
the positions where constraint violations are identified are used as
input for repair steps. This can be any position in the shift schedule. As
we will see in the next paragraph, in our problem the position in the
shift schedule will also determine which repair step is applied. In our
application, repair steps take the position as an input and then try
to find a possible modification while iterating through the shift schedule
beginning from a start position. The indication of a start position
allows some leeway in the application of the repair step. One
is not restricted concerning the start of the search for a successful
modification. The start position could be a random one, or the same
as the position of the violation, or whatever seems to be appropriate
for the problem at hand.

Taking a closer look at the roster of our shift scheduling problem
shown in Figure~\ref{figure:operationplan}, one can identify four main
clusters where operations might be the same within that cluster. From
Monday to Friday the operation is identical and there is only one
shift type involved. Since Thursday and Friday do have a more complex
operation plan, as they use different operation units, namely groups,
we have to split this cluster into two: One modification operator will
be defined for modifications from Mondays to Wednesdays, another one
will be defined for those from Thursdays to Fridays. The third
modification operator is focused on changes that involve the {\em day
shift at weekends\/} on Saturdays and Sundays. Finally, a repair step for
the {\em shift substitutions at weekends\/} on Saturdays is defined. In
the following one type of repair step will be explained.

\mbox{}\\
\noindent {\bf Repair step `Monday--Wednesday'}\\

The operation plan shows that on the days from Monday to Wednesday there is
only one shift type required, the {\em day shift}. If a violation of a
constraint and its derived repair start position is within this range,
there are two ways of applying a modification operator. 

\begin{figure}[th]
\epsfbox{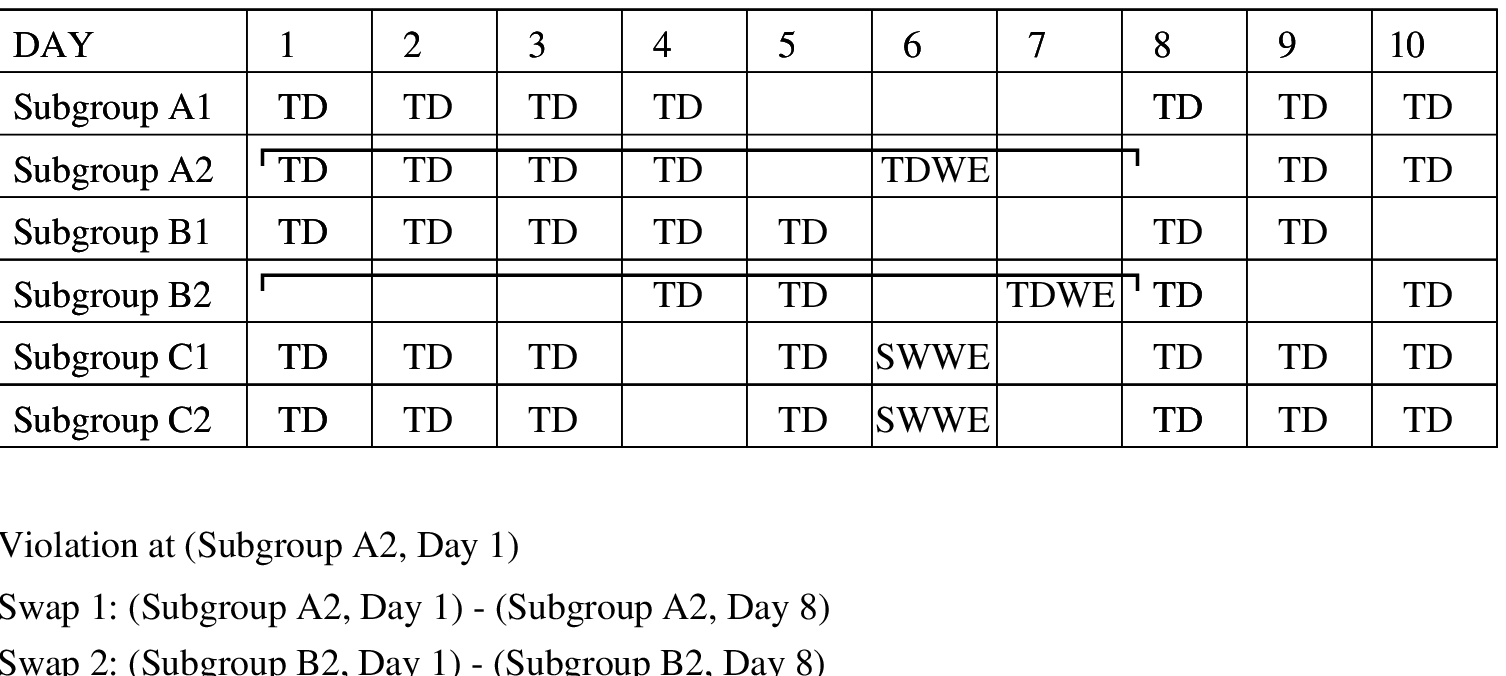}
\caption{Repair step 1 for Monday to Wednesday type of shifts.}
\label{figure:repairmonwed1}
\end{figure}

First, the position of the violation, which is characterized by the
day and the subgroup ($=$ smallest unit of operation in our problem)
affected, can be swapped with another position in the Monday to
Wednesday range. Actually, two swaps have to be made in order to preserve
the requirements of the operation plan. The reason is straightforward:
From Monday to Wednesday, five subgroups are allocated with {\em day
shifts}, resulting in one subgroup without a shift. If the violation is
on a position where a {\em day shift\/} is allocated, a swap is sought
with a position in the same subgroup where no shift is
allocated. Because the number of shifts allocated to subgroup is not
affected by such a swap this hard constraint is
preserved. Furthermore, by making a second opposite swap the hard
constraint for the required shifts at one day is preserved, too. Both
swaps involve an exchange of a free subgroup, i.e., one that has no
shift allocated, with a {\em day shift\/} subgroup. The operation is
illustrated in Figure~\ref{figure:repairmonwed1}.

\begin{figure}[th]
\epsfbox{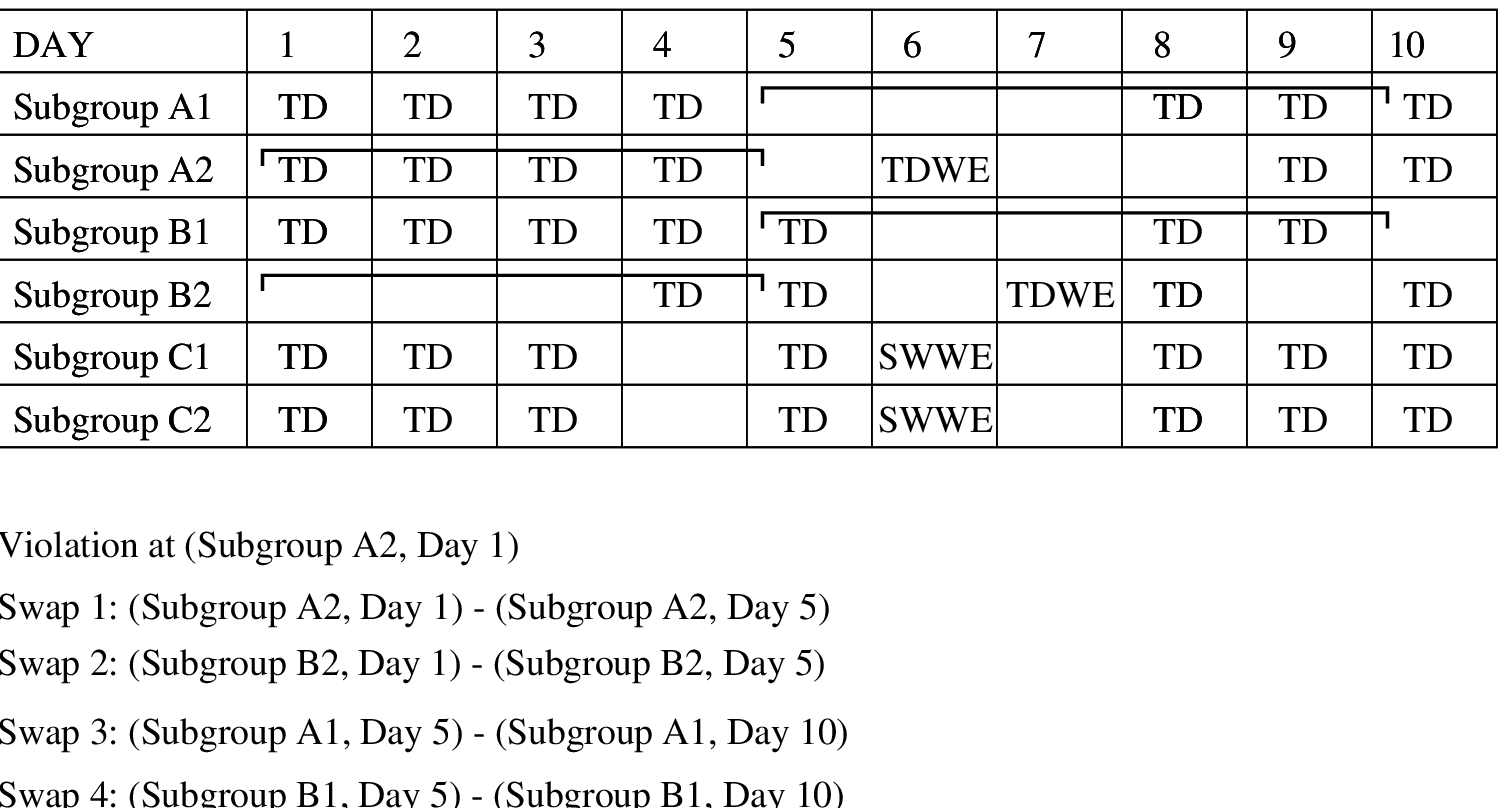}
\caption{Repair step 2 for Monday to Wednesday type of shifts.}
\label{figure:repairmonwed2}
\end{figure}

Second, since the {\em day shift\/} can also be found on Thursday and
Friday, it is also possible to seek destinations for swaps within this
range. There is only one thing that has to be considered in addition
to the above swap operation: On Thursday and Friday, groups are
required. Consequently, a swap operation must take care of this and
preserve the group structure of the operation requirements on Thursday
and Friday. This results in a more complex two-step modification
described in Figure~\ref{figure:repairmonwed2}.

\section{Results from the shift planning domain}\label{section:stats}

The algorithm that has been chosen out of the OptiFLIP++ library for
this problem is based on {\em iterative improvement techniques}. Such
an algorithm tries to improve an initial preliminary schedule
iteratively. The modifications, or repair steps, are the operations to
move from one schedule to another. The problem of getting trapped in a
local optimum can be overcome in several ways.

Our approach, which is a variant of the {\em iterative deepening}
heuristic described in Section~\ref{section:starflip}, works as
follows: In a first try, a search to depth 1 is made. This means that
starting from the initial schedule one modification is made. The
execution of one modification does imply that a certain number of
possible modifications of that type are tried on a random position out
of a set of the worst violated positions. After a number of such
tries, the best schedule survives as the new problem instantiation for
the next step and it is also remembered as the best overall
schedule. Hence, it is possible that one step produces a worse
intermediate schedule eliminating the possibility of getting trapped
in a local optimum. We discarded an earlier approach of avoiding local
optima traps that recursively increased the depth of a step if no
improvement could be achieved with current depth level due to
performance reasons.

Starting from the evaluated sets of constraints there are various
degrees of randomization that guide the optimizing algorithm. First,
the choice of which violated constraint to work on is randomized
within the set of worst violated constraints. Second, each type of
constraints has its own function of how to derive a position where the
repair modification is applied. If this function delivers several such
positions, one is randomly selected. Third, the set of possible
modifications applied by the repair operators provides another pool
for random choices.

\begin{figure}[th]
\epsfbox{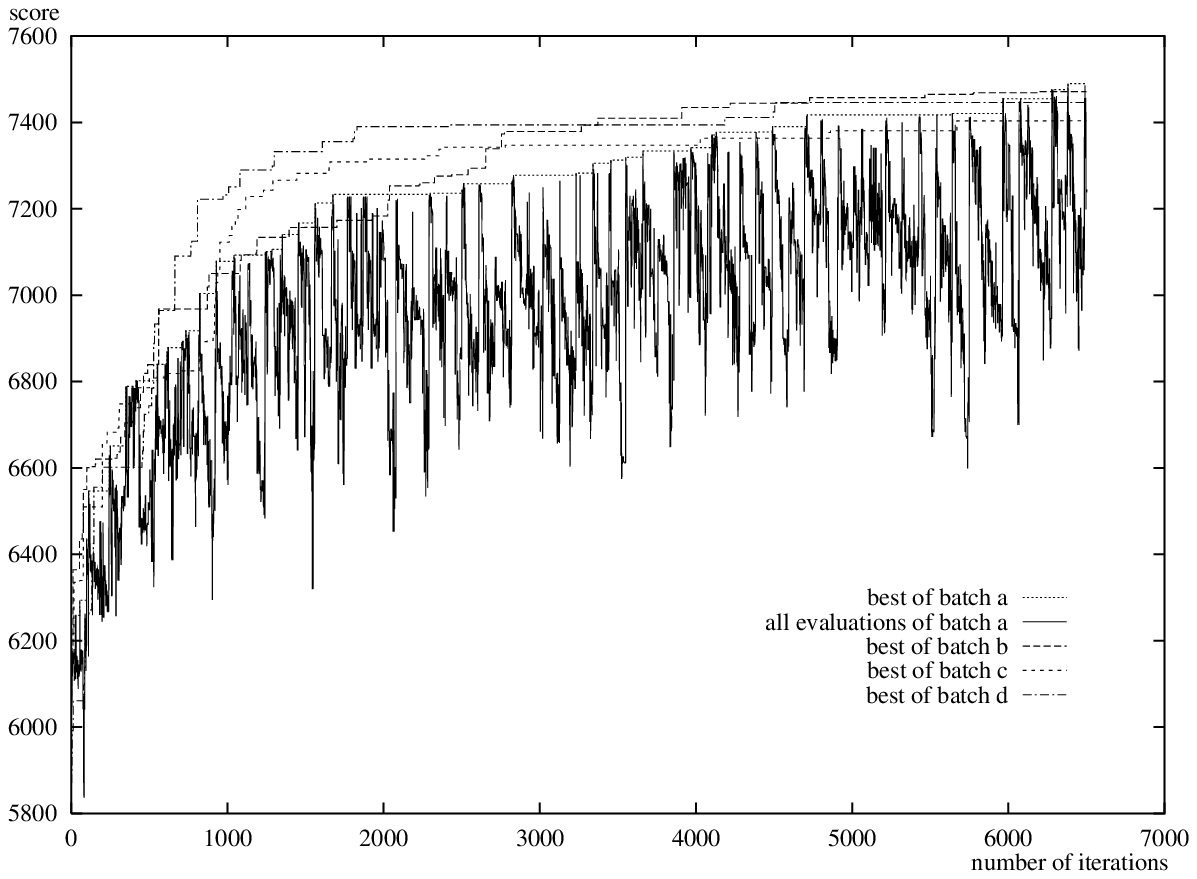}
\caption{Optimization results.}
\label{figure:data}
\end{figure}

In the following, we briefly summarize the results depicted in
Figure~\ref{figure:data} 
which empirically show the effectiveness of the StarFLIP++ libraries
in solving the shift scheduling problem as defined in
Section~\ref{section:shiftpb}. As one sees, results steadily improve
until further improvements can only be gained by unproportionally long
search sequences. The four depicted batches correspond to four
different iteration sequences starting with the same initial
suboptimal solution. Since the optimal curves do not differ very much,
we conclude that the presented optimization method is quite robust and
will usually be able to find adequate solutions.  In terms of
effective running time, `good' solutions with an objective function
above 7200 were found in the average after 1h44' on a Sparc-station 5
(170 MHz Turbo-SPARC processor) with 64 MB memory running under Solaris
2.5.1. A run with 6500 complete shift schedule evaluations took
8h15'. While these timings may seem large, it is no problem in the
shift scheduling context as there is more than enough time available
to optimize a schedule which will then be used for an extended period
of time. In reactive scheduling situations, simpler constraints and
faster hardware should make it possible to optimize the problem
efficiently. 


\section{Conclusions and outlook}

In a first expert system approach, Stohl et al.\ \cite{Stohl91}
applied a constructive domain heuristic to a steel making scheduling
problem. Although the system found good feasible solutions, Stohl et
al.\ believed that their solutions could be further improved,
especially since constraints could only be broken through explicit
user intervention, and because the relaxing of constraints was not
evaluated. The iterative optimization library StarFLIP++
allowed to handle these aspects and therefore proved to be more
suitable for the steel making scheduling problem.

In this paper, we reused enhanced StarFLIP++ components to present a
new shift scheduling problem as well as its solution.

This allows to highlight characteristics of major application areas
for StarFLIP++: Whenever rules can be elicited from human domain
experts, and when these rules are not absolute in the sense that they
can be more or less applicable for a certain data set, and
additionally one wants to allow trade offs to be made in order to find
adequate solutions, then this iterative optimization library to solve
combinatorial problems with approximate reasoning methods is well
suited for the problem at hand. It should be clear that these
characteristics apply to many industrial combinatorial optimization
problems, whereas artificially clean problems found in classical
operations research often do not fall in this category. However,
because of these characteristics, it is difficult to compare directly
the different methods as they do not solve the same kind of problems.

In this paper, we presented problems as well as solutions associated
with approximate reasoning methods in real world combinatorial
optimization problems. We presented the knowledge engineering tools
ConFLIP++ and DomFLIP++ for modeling fuzzy constraints that can be
aggregated to complex, hierarchical constraint structures. We showed
its practical application in a shift scheduling application using
fine-tuning and specializing concepts. We presented the DynaFLIP++
library which, based on ConFLIP++, is responsible for the evaluation
of an instantiated combinatorial optimization problem. We also gave an
overview of the heuristics and repair based OptiFLIP++ algorithms. We
developed a combination of repair based methods and fuzzy constraints
for real world multi criteria decision making, with a bias towards
scheduling problems. We presented improved methods for compromising
between antagonistic criteria, for assessing priorities among fuzzy
constraints, as well as a new method for ensuring consistent and
reasonable changes in configurations. We also introduced a method that
allows interactive what-if games for arbitrary decision problems. The
method is an argument based consistency test with a meta constraint
knowledge base that allows several experts to agree on parameters of a
knowledge base for real world decision making problems. Through the
consistency tested by the method, non-monotonic changes in knowledge
bases of combinatorial optimization problems can be made more
predictable.  Theoretical analysis and experiments indicate that our
method makes real world problems from this area manageable.

The results obtained from a shift scheduling application indicate the
suitability of our approach for similar combinatorial optimization
problems in terms of modeling expressiveness and performance.

Up to now all libraries have been implemented in the object oriented
language {\em C++}, which was the obvious choice at the start of the
project. Today, with more appealing programming languages and
object oriented concepts having reached a more mature and stable
level, there are other options available as far as the implementation
is concerned. In particular, the {\em JAVA\/} programming language
with such convenient standard features like networking classes
implying full Internet connectivity and a high degree of platform
independence is the first choice for future StarFLIP++
implementations. Nevertheless, {\em C++\/} is still a well justified
environment especially with such powerful extensions as the Standard
Template Library.

Current extensions aim at providing a distributed simulation package
over the Internet including an environment to test reactive scheduling
behavior. The programming of these extensions in Java instead of {\em
C++\/} should allow easier porting of the software to new computer
architectures, as well as help avoiding pitfalls encountered when
programming in {\em C++}.

\bibliographystyle{plain}
\bibliography{raggl}
\end{document}